\documentclass{bmvc2k}
\usepackage{gensymb}
\title{Structure-Aware 3D Hourglass Network for Hand Pose Estimation from Single Depth Image}

\addauthor{Fuyang Huang}{fyhuang@cse.cuhk.edu.hk}{1}
\addauthor{Ailing Zeng}{alzeng@cse.cuhk.edu.hk}{1}
\addauthor{Minhao Liu}{mhliu@cse.cuhk.edu.hk}{1}
\addauthor{Jing Qin}{harry.qin@polyu.edu.hk}{2}
\addauthor{Qiang Xu}{qxu@cse.cuhk.edu.hk}{1}

\addinstitution{
Computer Science \& Engineering\\
The Chinese University of Hong Kong\\
Hong Kong
}

\addinstitution{
School of Nursing\\
The Hong Kong Polytechnic University\\
Hong Kong
}

\def\etal{\emph{et al}\bmvaOneDot}
\runninghead{FUYANG \etal}{Structure-Aware Network for Hand Pose Estimation}

\begin{document}

\maketitle

\begin{abstract}
In this paper, we propose a novel structure-aware 3D hourglass network for hand pose estimation from a single depth image, which achieves state-of-the-art results on MSRA and NYU datasets. Compared to existing works that perform image-to-coordination regression, our network takes 3D voxel as input and directly regresses 3D heatmap for each joint. To be specific, we use hourglass network as our backbone network and modify it into 3D form. We explicitly model tree-like finger bone into the network as well as in the loss function in an end-to-end manner, in order to take the skeleton constraints into consideration. Final estimation can then be easily obtained from voxel density map with simple post-processing. Experimental results show that the proposed structure-aware 3D hourglass network is able to achieve a mean joint error of 7.4 mm in MSRA and 8.9 mm in NYU datasets, respectively. 
\end{abstract}

%-------------------------------------------------------------------------
\section{Introduction}
\label{sec:intro}
Accurate gesture recognition plays an important role in many applications (e.g., virtual reality and augmented reality). Articulated hand pose estimation, serving as a fundamental step towards gesture recognition, has thus drawn great attention from both industry and academia. 

Thanks to the availability of low-cost depth camera and recent advancements in machine learning techniques, many research efforts have been devoted to hand pose estimation from depth image in recent years. Despite the significant improvement in terms of accuracy and speed, current articulated hand pose estimation solution is still far from satisfactory. The main challenge lies in the fact that regressing from depth image to 3D coordination is a highly non-linear problem, which involves high degree of freedom (DOF) of hand pose, frequent self-occlusion, and background noises. 

In recent years, convolutional neural network (CNN) has been successfully applied in various types of computer vision tasks such as object detection, image classification, and human pose estimation~\cite{ShihEn2016,newell2016stacked,chu2017}. Similarly, discriminative data-driven approaches leveraging deep learning method overwhelm traditional generative model-driven approaches for hand pose estimation tasks in terms of accuracy and speed. It is regarded as a non-linear regression problem which regresses depth image to 3D joint coordination~\cite{tompson2014real,oberweger2015hands} and trained in a deep learning network.
% * <qxu@cse.cuhk.edu.hk> 2018-05-07T15:22:45.784Z:
% 
% The purpose of the above paragraph is rather vague
% 
% ^.

There are two main kinds of regression-based approaches. One directly regresses depth image to continuous joint positions~\cite{chen2017pose,oberweger2017deepprior++,oberweger2015hands,ge20173d,guo2017region} either in 2D or 3D space. The other outputs discrete probability heatmaps for each joint as a intermediate result and requires additional post-processing to get the final result. Recently, the heatmap-based approach has proved to be more promising in both human pose estimation \cite{Tompson2014,chu2017,ShihEn2016,Bulat2016,WYang2016,Pishchulin2016,Insafutdinov2016} and hand pose estimation \cite{tompson2014real,moon2018v2v,wan2018dense,ge2016robust}. \citeauthor{wan2018dense} \cite{wan2018dense} proposed a network that produces 2D heatmaps and 3D offset vector heatmaps to get the final 3D joint positions. It takes 2D depth image as input and employs 2D CNN to regress 3D heatmap. The depth data is originally captured in 3D space, but most of 2D CNN-based work uses projected 2D depth image to regress 3D coordination. Information loss is inevitable during the 3D-to-2D projection and 2D-to-3D regression. \citeauthor{moon2018v2v} \cite{moon2018v2v} proposed a voxel-to-voxel network with 3D ResNet \cite{he2016deep} as backbone network. They directly regress 3D heatmap of joint from 3D voxel to preserve as much information as possible,but they do not take skeleton constraints of hand pose into consideration in this network, which is essential for accurate hand pose estimation~\cite{tompson2014real,oberweger2017deepprior++,chen2017pose}. 
% * <qxu@cse.cuhk.edu.hk> 2018-05-07T15:33:23.431Z:
% 
% > However, the depth data is originally located in 3D space. The projected 2D depth image is not suitable for directly 3D regression due to the lack of 3D spatial information.
% This sentence is very ambiguous!
% 
% ^.

To tackle the above problems, we propose a novel structure-aware 3D hourglass network in this paper. Our framework is able to capture 3D spatial feature of depth data in all scale and embed skeleton constraints of hand pose in a single network. The input depth data and ground truth are first transformed into camera's coordination system and then discretized to voxels in 3D space. Subsequently, the voxels are regressed to 3D heatmaps of each joint, from which the final joint positions can be easily retrieved with simple post-processing. The regression model is trained by a stacked 3D hourglass network that captures all scales of 3D convolutional features in an end-to-end manner. In order to take skeleton constraints into consideration, we follow the tree structure of human hand and add additional channels at the final layer of each stack as bone heatmaps. These bone heatmaps work as intermediate supervision in the loss function and are then passed to next stack of hourglass network as input. Note that, the non-linear relationship between joint and skeleton structure can be learned implicitly in the following stacks of networks. Accordingly, the final joint heatmaps will be refined by the skeleton constraints of hand pose. To the best of our knowledge, this is the first work that regresses 3D depth data into 3D heatmaps with skeleton constraints  for 3D hand pose estimation. Experiments on two popular public datasets (MSRA~\cite{sun2015cascaded} and NYU~\cite{tompson2014real}) show that our method outperforms most of the state-of-the-art works in terms of mean joint error and success frame rate.  To sum up, the main contributions of this work include:

\vspace{-0.24cm}\begin{itemize}
\item This is the first work that incorporates hourglass network with depth image in purely 3D form to regress 3D joint heatmaps directly. Unlike most of previous hand pose estimation work leveraging hourglass network, our network consists of 3D residual network (ResNet) as basic building blocks and directly outputs 3D heatmap for each joint.
\item This is the first work that incorporates skeleton constraints of hand pose into detection network as intermediate supervision, which is different from most heatmap-based work that concatenates another refinement network or applies complex post-processing. This method is experimentally proved to be simple yet effective in hand pose estimation.
\end{itemize}

The remainder of this paper is organized as follows. In Section~\ref{sec:related}, literature on hand pose estimation is reviewed. In Section~\ref{sec:method}, architecture of our method is introduced. We show ablation study and comparison study with state-of-the-art work in Section~\ref{sec:exp}. Section~\ref{sec:conc} concludes this paper.
% From 2009, the proceedings of BMVC (the British Machine Visio
% Conference) will be published only in electronic form.  This document
% illustrates the required paper format, and includes guidelines on
% preparation of submissions.  Papers which fail to adhere to these
% requirements may be rejected at any stage in the review process.

% \LaTeX\ users should use this template in order to prepare their paper.
% Users of other packages should emulate the style and layout of this
% example.  Note that best results will be achieved using {\tt pdflatex},
% which is available in most modern distributions.

\section{Related Work}
\label{sec:related}
{\bf Generative Approach vs Discriminative Approach } Hand pose estimation from depth image has been extensively explored in recent years ~\cite{oberweger2017deepprior++,neverova2017hand,sharp2015accurate,wan2017crossing,guo2017region,moon2018v2v,wan2018dense,tagliasacchi2015robust}. There are two main streams of methods for this task, namely, model-driven generative approach and data-driven discriminative approach. The generative approach typically fits a pre-defined 3D hand model to a specific pose to match the input depth image ~\cite{sharp2015accurate,Dtang2015}, which can be regarded as an optimization problem. By shrinking the searching space, temporal information is usually involved in these optimizers such as Particle Swarm Optimizer (PSO) ~\cite{sharp2015accurate,sun2015cascaded} and Iterative Closest Point (ICP) ~\cite{tagliasacchi2015robust}. Despite high accuracy of generative methods, these methods highly rely on model initialization and temporal information, which lead to accumulative estimation error and re-initialization process in the running time. 

The discriminative approach has drawn more research attention these days~\cite{Keskin2012,Liang2014,sun2015cascaded,TangD2014,TangD2013,Dtang2015,CWan2016} to  directly localize hand joints from an input depth map. Accordingly, the literature regards the depth data either as a 2D depth image~\cite{ge2016robust,wan2018dense,guo2017region,tompson2014real,oberweger2017deepprior++} or as a 3D point cloud~\cite{ge20173d,moon2018v2v} in the pre-processing stage. Based on the data representation, the network structure of regression model can be categorized into 2D CNN-based and 3D CNN-based respectively. Considering the high non-linear relationship between input and output 3D joint positions, instead of regressing 3D joint position directly~\cite{oberweger2015hands,oberweger2017deepprior++,chen2017pose}, some of the researchers alternatively use heatmap to estimate final result~\cite{moon2018v2v,ge2016robust,wan2018dense,tompson2014real}

{\bf Direct Regression vs Joint Heatmap } For the direct regression model, a fully connected (FC) layer is appended to the end of the network to output continuous joint positions directly. \citeauthor{oberweger2017deepprior++} \cite{oberweger2017deepprior++} propose a framework introducing a bottleneck layer at the end of the network to direct regress 3D joint position while preserving skeleton constraints in the network. \citeauthor{chen2017pose} ~\cite{chen2017pose} propose a pose guided structured region ensemble network which is designed to capture tree-like structure of hand and is trained end-to-end in an iterative way. \citeauthor{tompson2014real} ~\cite{tompson2014real} is the first work that employs CNN to generate the joint heatmap.However, post-processing on 2D heatmap has an important limitation that self-occlusion of hand joints because they may share same depth value. To provide more data in different perspective, \citeauthor{ge2016robust} ~\cite{ge2016robust} project depth image into three different views and apply multi-view CNN to get multi-view heatmaps. Nevertheless, re-projecting 2D image to 3D space still underutilized spatial information of depth data.

{\bf 2D CNN vs 3D CNN} \citeauthor{wan2018dense} ~\cite{wan2018dense} use 2D CNN regress 2D heatmap and 3D heatmap at the same time, although there is information loss in the 2D-to-3D regression. \citeauthor{ge20173d} \cite{ge20173d} firstly employs 3D CNN to capture spatial feature in 3D space. \citeauthor{moon2018v2v} ~\cite{moon2018v2v} discretize 3D point cloud data into voxel grids and push them into a voxel-to-voxel network, which produces a set of 3D heatmaps for each joint. The network structure applies 3D ResNet as basic building block which is organized in a down-to-up structure and achieves state-of-the-art results in three public datasets. However, skeleton constraints of hand pose is not taken into consideration in this network which proved important in hand pose estimation system. 

According to aforementioned facts, we argue that 3D CNN is more suitable for 2.5D depth data because there is no information loss in the 3D-to-3D regression compared to 2D-to-3D regression. Meanwhile, 3D data presentation better preserves spatial information of depth data compared to the 2D image. Although more 3D CNN-based work has emerged recently, the power of 3D CNN is not fully investigated due to shallow network structure~\cite{ge20173d} or skeleton-unawareness~\cite{moon2018v2v}. So we propose a structure-aware 3D hourglass network to estimate hand pose. Our network takes 3D voxels as input and outputs 3D heatmap for each joint, employing 3D hourglass network as basic building block. Instead of performing IK-like optimization as post-processing, skeleton constraints of hand pose is treated as an intermediate supervision in the network and trained in an end-to-end manner. 
\begin{figure*}
\includegraphics[width=\linewidth]{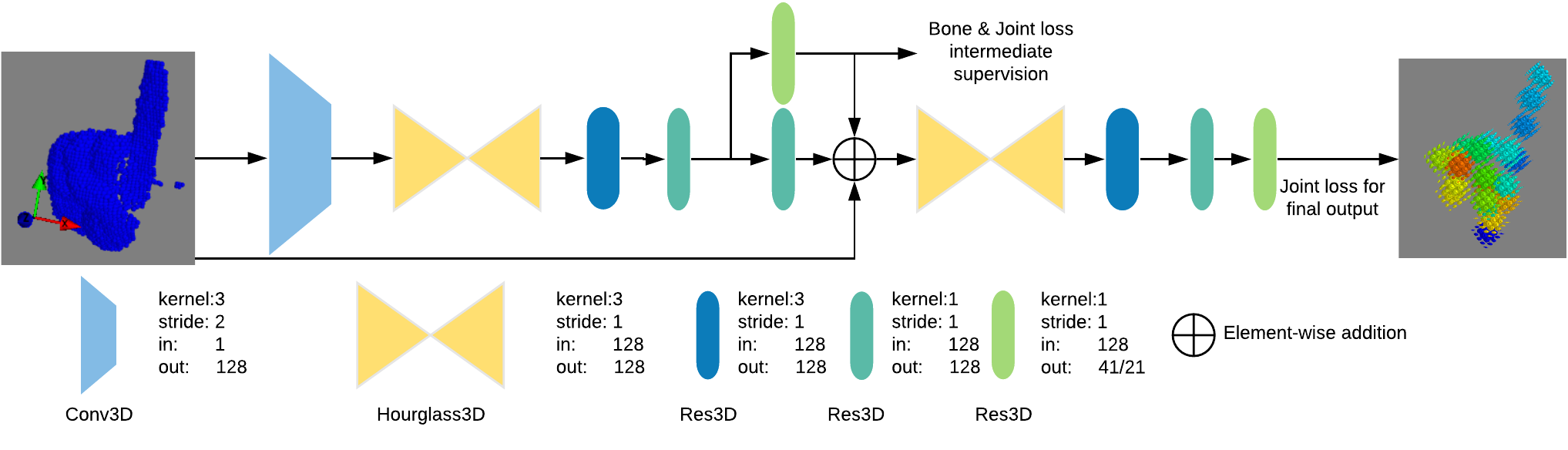}
\caption{To simplify the illustration, all the 3D modules are visualized with 2D shapes. The two-stacks 3D hourglass network starts with discretized 3D volumetric data. The volume data is first down-sampled to $32\times32\times32$ by a $3\times3$ 3D CNN layer with stride 2. After feature maps propagates through the first hourglass module and Res3D module, bone and joint heatmaps are generated by two consecutive $1\times1$ Res3D modules as intermediate supervision. The heatmaps, together with original data and feature maps, are fed into next stack for further propagation.}
\label{fig:overall}
\end{figure*}

\section{Proposed Method}
\label{sec:method}
We define the 3D hand pose estimation from depth image as a voxel-wise regression problem. Our method takes normalized binary voxels as input and outputs 3D heatmap of hand joints in voxel space. The input depth image is first reprojected to camera space as point cloud data. Then, the point cloud data is discretized into binary voxels where normalization is performed according to the size of point cloud and then the binary voxels are passed to the network for forward propagation. The details of the proposed network is described in section~\ref{sec:net}. The training target is 3D Gaussian distribution of joint probability in voxel form generated from ground truth. To involve in the skeleton constraints of hand pose, 3D bone probability heatmaps are trained as an intermediate supervision in the network simultaneously. The overall framework is shown in Fig~\ref{fig:overall}.
\vspace{-0.35cm}
\subsection{Pre-processing}
Before being fed into a 3D hourglass network, input data should be transformed in volumetric representation. The depth images are first re-projected to camera space by camera's intrinsic parameter. In order to truncate the point cloud into a cubic box, we simply regard the geometric center of points as box center and maximum length along x-,y-, and z-axis as box length. Afterward, the point cloud will be discretized into binary voxel grids where voxel with value one indicating it is occupied by depth data and zero otherwise. As for the ground truth, they are also transformed into voxel grids. However, the ground truth heatmap for one joint only contains one positive point which is too sparse for training. So we perform 3D Gaussian distribution on the ground truth voxels whose center is located at the ground truth point with the standard deviation of voxel grid length. A sample heatmap image is also shown in Fig~\ref{fig:overall}.
\vspace{-0.1cm}
\subsection{Network Structure}
\label{sec:net}
The basic building block of our network is hourglass network proposed by \citeauthor{newell2016stacked} ~\cite{newell2016stacked} in 2016 which is widely used in human pose estimation. The bottom-up and top-down scheme of hourglass network is effective in dense prediction tasks such as semantic segmentation and heatmap prediction. Additionally, a set of hourglass networks can be stacked together to employ intermediate supervision where we can involve in skeleton constraints of hand. 

Different from ~\cite{newell2016stacked}, our network takes 3D volumetric data as input. Thus, 3D CNN layers consist the basic module of the network. Considering the large consumption of GPU memory of 3D CNN, we set the input resolution of voxel to $64\times64\times64$, output resolution to $32\times32\times32$ and stack number to 2. Before being fed into hourglass module, the input voxels are down-sampled by a $3\times3\times3$ CNN layer with stride 2. The number of channels of each residual block in hourglass is 128. After forwarding through each hourglass module, the voxel-wise heatmaps of $J$ joints are predicted by two consecutive $1\times1\times1$ CNN layers at the end of each stack. 

\subsection{Training Target}
To predict voxel-wise heatmap for each joint, we need to generate training target first. Since there is only one ground truth value for each joint in a single 3D heatmap, direct regression may lead to overfitting. Similar to many dense prediction work before ~\cite{tompson2014real,newell2016stacked}, we perform 3D Gaussian distribution centered at the ground truth position of each joint with standard deviation of voxel length, which is then compared to the predicted heatmap by Mean-Squared Error (MSE) loss. Specially, the loss for joints is defined in equation~\ref{equa:l1}:
\begin{equation}
\label{equa:l1}
L_j = \sum_{s=1}^{S}\sum_{n=1}^{J}\sum_{i,j,k}^{R}|H_{s,n}^{j}(i,j,k)-\hat{H_{n}^{j}}(i,j,k)|^{2}
\end{equation}
where $H_{s,n}^{j}(i,j,k)$ indicates the predicted value at $i,j,k$ of heatmap for $n$th joint in $s$th stack. $\hat{H_{n}^{j}}$ is the ground truth heatmap for $n$th joint. $S,J$ and $R$ denote stack number, joint number and resolution respectively. 

\subsection{Skeleton constraints}
As discussed before, explicitly leveraging skeleton constraints of hand helps to increase prediction accuracy in hand pose estimation task~\cite{oberweger2017deepprior++,chen2017pose}. However, heatmap-based hand pose estimation tasks either let the skeleton constraints trained implicitly by the network~\cite{wan2018dense,moon2018v2v} or define it as an additional optimization problem in the post processing stage~\cite{ge2016robust}. In our proposed network, besides heatmap for each joint, heatmaps for each bone are also learned in the network as an intermediate supervision. The number of bones are determined by the tree structure of human hand. For example, in MSRA dataset, there are 21 joints in each sample, connected by 20 bones. The bone heatmaps are generated in the similar way as joint heatmap with smaller standard deviation which equals to 0.5 times of voxel length. Additionally, bone-to-joint relationship can be learned in our stacked network, benefiting joint refinement in the final stack. In order to learn the bone-to-joint relationship, bone heatmaps are supervised at the end of each stack except for the last stack, as shown in Fig~\ref{fig:overall}. The bone loss as intermediate supervision is defined in equation~\ref{equa:l2}.
\begin{equation}
\label{equa:l2}
L_b = \sum_{s=1}^{S-1}\sum_{n=1}^{B}\sum_{i,j,k}^{R}|H_{s,n}^{b}(i,j,k)-\hat{H_{n}^{b}}(i,j,k)|^{2}
\end{equation}
where $H_{s,n}^{b}(i,j,k)$ represents the predicted value at $i,j,k$ of heatmap for $n$th bone in $s$th stack and B denotes the total number of bones.

Thus, the final loss function for the entire network is defined as $L = L_{j}+L_{b}$
\subsection{Post processing}
Because the average length of one voxel with $32\times32\times32$ resolution is 8mm and if we simply choose the top responding voxel as final result, the error caused by discretization could be as much as $\sqrt[]{4^2+4^2+4^2} = 6.92$mm. Hence, we adopt a simple strategy to recover joint position as shown in equation~\ref{equa:l4}:

\begin{equation}
\label{equa:l4}
  \begin{aligned}
  J_n &= \sum^K_{i=1}{w_i^{(n)}j_i^{(n)}} \\
  \text{where}~w_i^{(n)} &= \frac{H_{n}^{j}(j_i^{(n)})}{\sum^K_{i=1}{H_{n}^{j}(j_i^{(n)})}}
  \end{aligned}
\end{equation}

For $n$th joint, the final position is the weighted mean of top $K$ responding voxel of heatmap $n$. $j_i^{(n)}$ denotes the position of top $i$ responding voxel for joint $n$ and $w_i^{(n)}$ denotes the weight of each voxel whose sum is normalize to 1 by the corresponding value $H_{n}^{j}(j_i^{(n)})$ of predicted joint heatmap $n$. This strategy contributes 0.3mm accuracy to final result in MSRA dataset.

\section{Experimental Results}
\label{sec:exp}
Our method is tested on two public datasets \textemdash MSRA ~\cite{sun2015cascaded} and NYU ~\cite{tompson2014real}. We measure the mean error distance in 3D space and percentage of success frame (error of all joints in a single frame are below a threshold) as our measuring metrics. Hand region is cropped out in MSRA dataset, while NYU dataset retains the original depth data including background which requires additional hand detection step. To test our hand pose estimation network in a fair way, we use MSRA dataset in the ablation study to eliminate the influence of hand detection.
\subsection{Datasets}
{\bf MSRA Hand Pose Dataset.} There are 75K images in MSRA dataset consisting of 17 gestures from 9 subjects~\cite{sun2015cascaded}. Hand region is cropped out in this dataset, but testing set is not specified. Similar to what most previous work do~\cite{wan2017crossing,moon2018v2v}, we apply leave-one-out cross-validation strategy to split the dataset into training set with 8 subjects and testing set with 1 subject. 

{\bf NYU Hand Pose Dataset.} There are 72k images for training and 8k images for testing in NYU dataset~\cite{tompson2014real} whose annotations of hand pose contain 36 joints.Because hand region is not cropped out in this dataset, we directly crop the hand region in a cubic box whose center is located at center-of-mass of hand from ground truth. In our experiment,we also followed most of the previous works using frames from the frontal view and 14 out of 36 joints in the evaluation.

\subsection{Experiment Setting}
In order to facilitate different orientation and aspect ratio of input data, we perform data augmentation, which proved to be an effective approach in discriminative approach-based hand pose estimation task~\cite{oberweger2017deepprior++}. Specifically, we implement 3D data augmentation by rotating point cloud and by changing the aspect ratio in $xy$ space of the 3D space. Each training sample is randomly rotated from $-30\degree$ to $30\degree$ and the aspect ratio is scaled from $0.8$ to $1.2$ during training.

We run RMSProp optimizer in the training stage. All weights are initialized randomly from scratch.The input and output resolution is $64\times64\times64$ and $32\times32\times32$ respectively. The learning rate is initially set to 1e-5 and decays 0.3 every 5 epochs. Our system is implemented by PyTorch and trained on a single NVIDIA TITAN X GPU with batch size 16 and epoch number 20.

\subsection{Ablation Study}
\label{sec:ab}
In the ablation study, we aim to answer three questions: Does heatmap-based regression outperforms direct coordination regression? Does 3D data representation better interprets 2.5D depth data over 2D representation in hand pose estimation task?  Does skeleton constraints term help to improve accuracy? 

\begin{table}

\begin{center}
\begin{tabular}{|l|c|}
\hline
Baselines & Mean Joint Error  \\
\hline\hline
3D Direct regression without skeleton loss (B1) & $11.6$mm \\
3D Heatmap without skeleton loss           (B2) & $7.9$mm\\
2D Direct regression without skeleton loss (B3) & $13.1$mm \\
2D heatmap without skeleton loss           (B4) & $8.1^*$mm \\
2D heatmap with skeleton loss              (B5) & $7.6^*$mm\\
3D heatmap with skeleton loss        (Proposed) & $7.4$mm\\
\hline
\end{tabular}
\end{center}
\caption{Mean joint error results for five baselines and proposed method on MSRA dataset. B is short for baseline. $*$ indicates that error is only measured in $xy$ space.}
\label{tab:self}
\end{table}

{\bf Direct Regression \& Heatmap. }For the first experiment, we want to find whether heatmap-based regression outperforms direct coordination direction. So the first baseline shares the same network basic building block (3D hourglass) with the our proposed network except for some changes in the end of network. For baseline 1 direct regression, two consecutive fully connected layers replace original $1\times1$ convolutional layer to direct output 3D joint positions without skeleton constraint loss. For baseline 2, we only remove the skeleton constraints in the output layer compared to our proposed network. By comparing the result of baseline 1 and 2, direct coordination regression performs inferior to heatmap-based method with regard to mean joint error as shown in Table \ref{tab:self}. 

{\bf 2D \& 3D Data Representation. } For the second experiment, we evaluate the impact of 2D and 3D data representation on estimation. Here we introduce baseline 3 which takes 2D depth image as input and consists of 2D hourglass modules. The output layer is the same as that of baseline 1 network which regresses 3D coordination directly. As can be seen in Table ~\ref{tab:self}, 3D data representation outperforms 2D representation by a large margin. Meanwhile, it is more compute-intensive.because of 3D convolutional operation.

\begin{figure}
\begin{center}
\includegraphics[width=0.5\linewidth]{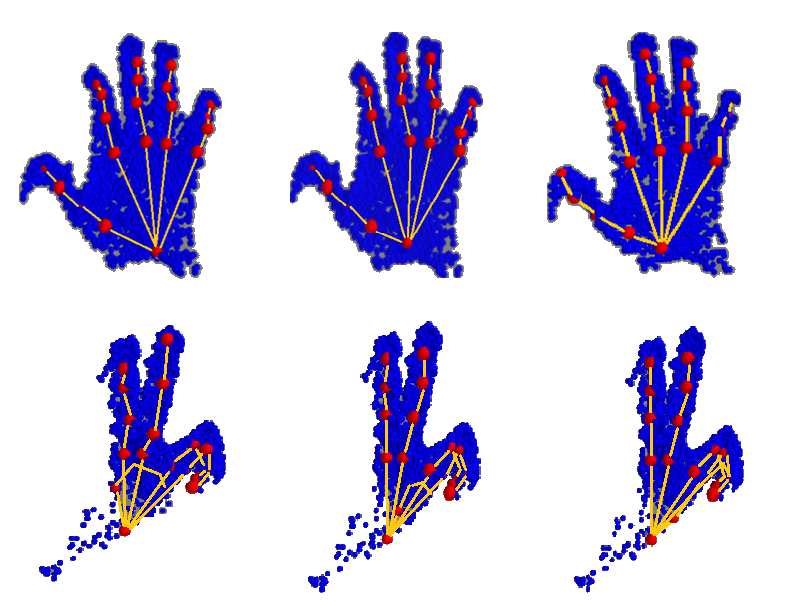}
\end{center}
\caption{Qualitative results of Baseline 2 (left), proposed method (middle) and ground truth (right) on MSRA dataset, visualized in voxel space.}
\label{fig:qual}
\end{figure}

{\bf Skeleton constraints.  } For the last experiment, we evaluate the effectiveness of bone loss in our network quantitatively and qualitatively. For quantitative comparison, a straight way is to compare the result of baseline 2 and proposed method which indicates that skeleton constraints reduces the mean error by 1mm. To further prove the effectiveness of bone loss, we conduct baseline 4 and baseline 5 which apply 2D hourglass network and output 2D heatmap. Baseline 5 involves in bone heatmap while baseline 4 does not. Because it is 2D heatmap, we only measure the joint error in $xy$ space. The performance of baseline 5 is also better than baseline 4. We additionally conduct qualitative comparison between baseline 2 and proposed method. As shown in Fig~\ref{fig:qual}, compared to baseline 2, result of proposed method is more realistic. To be specific, as shown in the second row, finger bones are constrained by skeleton model, showing a more straight bone and a more reasonable bone length than baseline 2 result. We can conclude that the bone heatmap helps in hand pose estimation. 

\begin{table}[ht]
\begin{minipage}{0.48\linewidth}
\begin{tabular}{|l|c|}
\hline
Method (MSRA) & Mean Error\\
\hline\hline
Multiview-CNN ~\cite{ge2016robust}  & 13.2mm \\ 
Deepprior++ ~\cite{oberweger2017deepprior++} & 9.5mm \\
3DCNN ~\cite{ge20173d} & 9.5mm \\
Crossing Net ~\cite{wan2017crossing} &  12.2mm\\ 
Cascaded ~\cite{sun2015cascaded} & 15.2mm \\ 
REN ~\cite{guo2017region} &  9.7mm \\
Pose-REN ~\cite{chen2017pose} &  8.65mm\\ 
V2V-net ~\cite{moon2018v2v} & 7.59mm \\ 
Dense ~\cite{wan2018dense} & \textbf{7.2mm} \\ 
3D hourglass (Ours) & 7.4mm\\ 
\hline
\end{tabular}
\end{minipage}
\begin{minipage}{0.48\linewidth}
\begin{tabular}{|l|c|}
\hline
Method (NYU) & Mean Error \\
\hline\hline
Deepprior++ ~\cite{oberweger2017deepprior++} & 12.24mm \\
3DCNN ~\cite{ge20173d} & 14.1mm \\
Crossing Net ~\cite{wan2017crossing} & 15.5mm \\
V2V-net ~\cite{moon2018v2v} & \textbf{8.42mm} \\
REN ~\cite{guo2017region} &  12.69mm\\
Pose-REN ~\cite{chen2017pose} & 11.81mm \\
Dense ~\cite{wan2018dense}& 10.2mm \\
3D hourglass (Ours) & 8.9mm\\
\hline
\end{tabular}
\end{minipage}
\vspace{+0.2cm} 
\caption{Comparison result regarding mean joint error on MSRA and NYU dataset.}
\label{tab:comp}
\end{table}

\begin{figure}[ht]
\begin{center}
\includegraphics[width=0.8\linewidth]{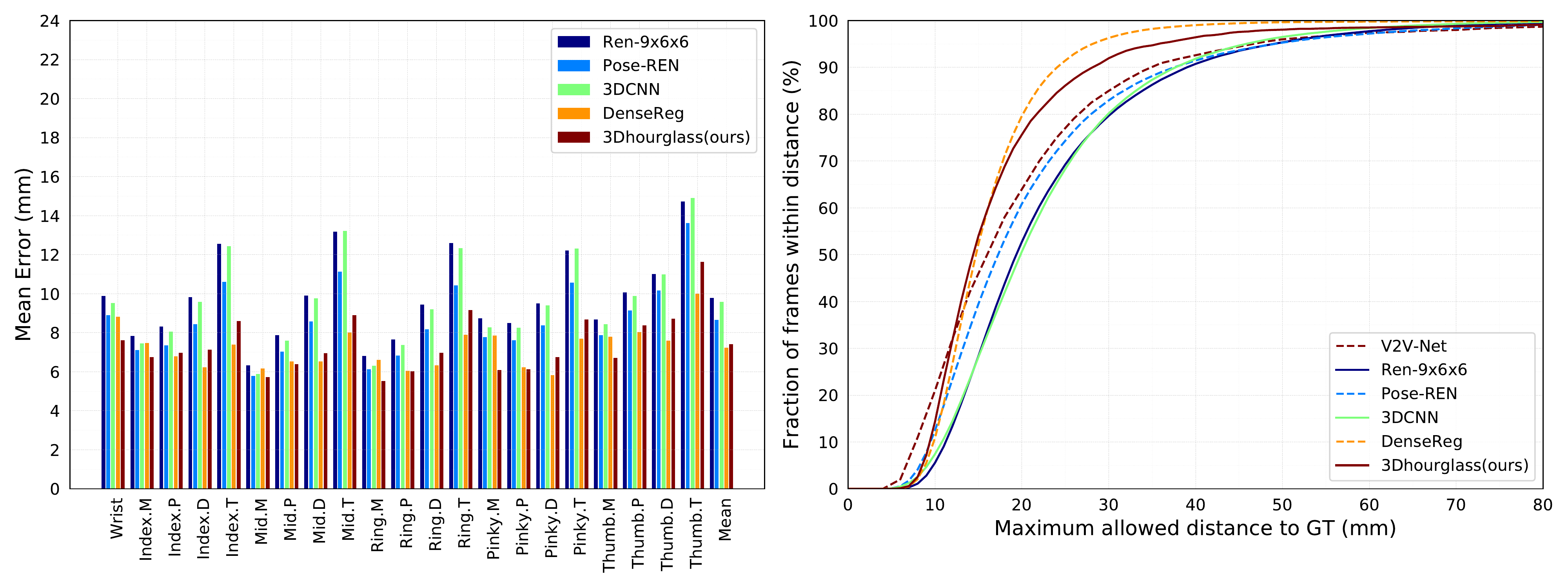}
\end{center}
\caption{Result of per-joint mean error (left) and success frame rate (right) on MSRA dataset}
\label{fig:result}
\end{figure}

\subsection{Comparison to State-of-the-art Solutions}
We compare our proposed network to several state-of-the-art hand pose estimation work on MSRA and NYU dataset~\cite{wan2018dense,chen2017pose,guo2017region,ge20173d,wan2017crossing,oberweger2017deepprior++,sun2015cascaded,moon2018v2v,ge2016robust}. Specifically, V2V-net ~\cite{moon2018v2v} and 3DCNN ~\cite{ge20173d} are 3D CNN-based work. Cascaded regression (Cascaded) ~\cite{sun2015cascaded}, region ensemble network (REN) ~\cite{guo2017region}, pose-guided regional ensemble network (Pose-REN) ~\cite{chen2017pose}, Deepprior++ ~\cite{oberweger2017deepprior++} and GAN-based Crossing Net ~\cite{wan2017crossing} are direct regression work. Dense regression model (DENSE)~\cite{wan2018dense} and Multiview-CNN ~\cite{ge2016robust} are heatmap-based work. 

As shown in Fig~\ref{fig:result} and Table~\ref{tab:comp}, our work outperforms most of the state-of-the-art works in terms of mean joint error and success frame. For mean joint error, our result ranks second in both of MSRA dataset (7.4mm) and NYU dataset (8.9mm) separately, indicating its advantage of cross-dataset stability. For the success frame, our work achieves the best result on MSRA dataset when threshold is around 15mm and the second best of overall performance.

As can be seen from Table~\ref{tab:comp}, ~\cite{wan2018dense} and ~\cite{moon2018v2v} show best performance on MSRA and NYU respectively. Compared to~\cite{wan2018dense}, our method achieves a very close result on MSRA dataset and a much better result on NYU dataset. The main reason is that NYU dataset is captured by structure light camera which shows many defective pixels (black holes) on the depth image. These black holes significantly influence the convolutional operation in 2D space because of large responsive value at edge, while 3D CNN is tolerative to these defective pixels due to the binary data representation in voxel space. Meanwhile, our method outperforms another 3D CNN-based V2V-Net~\cite{moon2018v2v} on MSRA dataset due to the skeleton constraint which is proved effective in section~\ref{sec:ab}. Additionally, the running speed of our system during evaluation is about 10.8 fps compared to 3.5 fps in \cite{moon2018v2v} with the same GPU type.
 The qualitative results are shown in Fig~\ref{fig:qual}.

\section{Conclusion}
\label{sec:conc}
In this paper, we propose a structure-aware 3D hourglass network to estimate hand pose from single depth image. To fully utilize the 2.5D depth data, we re-project it into 3D space and regress to 3D joint heatmap by 3D hourglass network. Despite joint loss, bone loss is introduced into the network as intermediate supervision to explicitly model skeleton constraints. Experimental results show that 3D data representation, 3D heatmap and skeleton constraint all contribute a lot to the final result. And our result on two datasets is on par with state-of-the-art result.

\vspace{0.2cm}
\hspace{-0.6cm}
\textbf{Acknowledgement:} This project was funded by National Natural Science Foundation of China (Grant No. 61432017) and Innovation and Technology Support Programme (Project Ref. ITS/360/16).
\bibliography{egbib}

\begin{thebibliography}{29}
\providecommand{\natexlab}[1]{#1}
\providecommand{\url}[1]{\texttt{#1}}
\expandafter\ifx\csname urlstyle\endcsname\relax
  \providecommand{\doi}[1]{doi: #1}\else
  \providecommand{\doi}{doi: \begingroup \urlstyle{rm}\Url}\fi

\bibitem[Bulat(2016)]{Bulat2016}
Tzimiropoulos~G Bulat, A.
\newblock Human pose estimation via convolutional part heatmap regression.
\newblock In \emph{European Conference on Computer Vision}, pages 717--732,
  2016.

\bibitem[C.~Keskin and Akarun(2012)]{Keskin2012}
Y.~E.~Kara C.~Keskin, F.~Kırac¸ and L.~Akarun.
\newblock Hand pose estimation and hand shape classification using
  multi-layered randomized decision forests.
\newblock In \emph{In European Conference on Computer Vision}, pages 852--863,
  2012.

\bibitem[C.~Wan and Gool(2016)]{CWan2016}
A.~Yao C.~Wan and L.~Van Gool.
\newblock Hand pose estimation from local surface normals.
\newblock In \emph{In European Conference on Computer Vision}, pages 554--569,
  2016.

\bibitem[Chen et~al.(2017)Chen, Wang, Guo, and Zhang]{chen2017pose}
Xinghao Chen, Guijin Wang, Hengkai Guo, and Cairong Zhang.
\newblock Pose guided structured region ensemble network for cascaded hand pose
  estimation.
\newblock \emph{arXiv preprint arXiv:1708.03416}, 2017.

\bibitem[Chu(2017)]{chu2017}
Yang W. Ouyang W. Ma C. Yuille-A.L. Wang~X Chu, X.
\newblock Multicontext attention for human pose estimation.
\newblock In \emph{2017 IEEE Conference on Computer Vision and Pattern
  Recognition (CVPR)}, pages 5669--5678, 2017.

\bibitem[D.~Tang and Kim(2014)]{TangD2014}
A.~Tejani D.~Tang, H. Jin~Chang and T.-K. Kim.
\newblock Latent regression forest: Structured estimation of 3d articulated
  hand posture.
\newblock In \emph{In IEEE Conference on Computer Vision and Pattern
  Recognition}, pages 3786--3793, 2014.

\bibitem[D.~Tang and Shotton(2015)]{Dtang2015}
P.~Kohli C. Keskin T.-K.~Kim D.~Tang, J.~Taylor and J.~Shotton.
\newblock Opening the black box: Hierarchical sampling optimization for
  estimating human hand pose.
\newblock In \emph{In IEEE International Conference on Computer Vision}, pages
  3325--3333, 2015.

\bibitem[D.~Tang and Kim(2013)]{TangD2013}
T.-H.~Yu D.~Tang and T.-K. Kim.
\newblock Real-time articulated hand pose estimation using semi-supervised
  transductive regression forests.
\newblock In \emph{In IEEE International Conference on Computer Vision}, pages
  3224--3231, 2013.

\bibitem[E.~Insafutdinov and Schiele(2016)]{Insafutdinov2016}
B.~Andres M.~Andriluka E.~Insafutdinov, L.~Pishchulin and B.~Schiele.
\newblock Deepercut: A deeper, stronger, and faster multiperson pose estimation
  model.
\newblock In \emph{European Conference on Computer Vision}, pages 34--50, 2016.

\bibitem[Ge et~al.(2016)Ge, Liang, Yuan, and Thalmann]{ge2016robust}
Liuhao Ge, Hui Liang, Junsong Yuan, and Daniel Thalmann.
\newblock Robust 3d hand pose estimation in single depth images: from
  single-view cnn to multi-view cnns.
\newblock In \emph{Proceedings of the IEEE Conference on Computer Vision and
  Pattern Recognition}, pages 3593--3601, 2016.

\bibitem[Ge et~al.(2017)Ge, Liang, Yuan, and Thalmann]{ge20173d}
Liuhao Ge, Hui Liang, Junsong Yuan, and Daniel Thalmann.
\newblock 3d convolutional neural networks for efficient and robust hand pose
  estimation from single depth images.
\newblock In \emph{Proceedings of the IEEE Conference on Computer Vision and
  Pattern Recognition}, volume~1, page~5, 2017.

\bibitem[Guo et~al.(2017)Guo, Wang, Chen, Zhang, Qiao, and Yang]{guo2017region}
Hengkai Guo, Guijin Wang, Xinghao Chen, Cairong Zhang, Fei Qiao, and Huazhong
  Yang.
\newblock Region ensemble network: Improving convolutional network for hand
  pose estimation.
\newblock \emph{arXiv preprint arXiv:1702.02447}, 2017.

\bibitem[H.~Liang and Thalmann(2014)]{Liang2014}
J.~Yuan H.~Liang and D.~Thalmann.
\newblock Parsing the hand in depth images.
\newblock pages 1241--1253, 2014.

\bibitem[He et~al.(2016)He, Zhang, Ren, and Sun]{he2016deep}
Kaiming He, Xiangyu Zhang, Shaoqing Ren, and Jian Sun.
\newblock Deep residual learning for image recognition.
\newblock In \emph{Proceedings of the IEEE conference on computer vision and
  pattern recognition}, pages 770--778, 2016.

\bibitem[L.~Pishchulin and Schiele.(2016)]{Pishchulin2016}
S.~Tang B. Andres M. Andriluka P. V.~Gehler L.~Pishchulin, E.~Insafutdinov and
  B.~Schiele.
\newblock Deepcut: joint subset partition and labeling for multi person pose
  estimation.
\newblock In \emph{2016 IEEE Conference on Computer Vision and Pattern
  Recognition (CVPR)}, pages 4929--4937, 2016.

\bibitem[Moon et~al.(2018)Moon, Chang, and Lee]{moon2018v2v}
Gyeongsik Moon, Ju~Yong Chang, and Kyoung~Mu Lee.
\newblock V2v-posenet: Voxel-to-voxel prediction network for accurate 3d hand
  and human pose estimation from a single depth map.
\newblock In \emph{CVPR}, volume~2, 2018.

\bibitem[Neverova et~al.(2017)Neverova, Wolf, Nebout, and
  Taylor]{neverova2017hand}
Natalia Neverova, Christian Wolf, Florian Nebout, and Graham~W Taylor.
\newblock Hand pose estimation through semi-supervised and weakly-supervised
  learning.
\newblock \emph{Computer Vision and Image Understanding}, 164:\penalty0 56--67,
  2017.

\bibitem[Newell et~al.(2016)Newell, Yang, and Deng]{newell2016stacked}
Alejandro Newell, Kaiyu Yang, and Jia Deng.
\newblock Stacked hourglass networks for human pose estimation.
\newblock In \emph{European Conference on Computer Vision}, pages 483--499.
  Springer, 2016.

\bibitem[Oberweger and Lepetit(2017)]{oberweger2017deepprior++}
Markus Oberweger and Vincent Lepetit.
\newblock Deepprior++: Improving fast and accurate 3d hand pose estimation.
\newblock In \emph{ICCV workshop}, volume 840, page~2, 2017.

\bibitem[Oberweger et~al.(2015)Oberweger, Wohlhart, and
  Lepetit]{oberweger2015hands}
Markus Oberweger, Paul Wohlhart, and Vincent Lepetit.
\newblock Hands deep in deep learning for hand pose estimation.
\newblock \emph{arXiv preprint arXiv:1502.06807}, 2015.

\bibitem[Sharp et~al.(2015)Sharp, Keskin, Robertson, Taylor, Shotton, Kim,
  Rhemann, Leichter, Vinnikov, Wei, et~al.]{sharp2015accurate}
Toby Sharp, Cem Keskin, Duncan Robertson, Jonathan Taylor, Jamie Shotton, David
  Kim, Christoph Rhemann, Ido Leichter, Alon Vinnikov, Yichen Wei, et~al.
\newblock Accurate, robust, and flexible real-time hand tracking.
\newblock In \emph{Proceedings of the 33rd Annual ACM Conference on Human
  Factors in Computing Systems}, pages 3633--3642. ACM, 2015.

\bibitem[Sun et~al.(2015)Sun, Wei, Liang, Tang, and Sun]{sun2015cascaded}
Xiao Sun, Yichen Wei, Shuang Liang, Xiaoou Tang, and Jian Sun.
\newblock Cascaded hand pose regression.
\newblock In \emph{Proceedings of the IEEE Conference on Computer Vision and
  Pattern Recognition}, pages 824--832, 2015.

\bibitem[Tagliasacchi et~al.(2015)Tagliasacchi, Schr{\"o}der, Tkach, Bouaziz,
  Botsch, and Pauly]{tagliasacchi2015robust}
Andrea Tagliasacchi, Matthias Schr{\"o}der, Anastasia Tkach, Sofien Bouaziz,
  Mario Botsch, and Mark Pauly.
\newblock Robust articulated-icp for real-time hand tracking.
\newblock In \emph{Computer Graphics Forum}, volume~34, pages 101--114. Wiley
  Online Library, 2015.

\bibitem[Tompson(2014)]{Tompson2014}
Arjun; LeCun Yann; Bregler~Christoph Tompson, Jonathan;~Jain.
\newblock Joint training of a convolutional network and a graphical model for
  human pose estimation.
\newblock In \emph{Advances in Neural Information Processing Systems}, pages
  1799--1807, 2014.

\bibitem[Tompson et~al.(2014)Tompson, Stein, Lecun, and
  Perlin]{tompson2014real}
Jonathan Tompson, Murphy Stein, Yann Lecun, and Ken Perlin.
\newblock Real-time continuous pose recovery of human hands using convolutional
  networks.
\newblock \emph{ACM Transactions on Graphics (ToG)}, 33\penalty0 (5):\penalty0
  169, 2014.

\bibitem[Wan et~al.(2017)Wan, Probst, Van~Gool, and Yao]{wan2017crossing}
Chengde Wan, Thomas Probst, Luc Van~Gool, and Angela Yao.
\newblock Crossing nets: Combining gans and vaes with a shared latent space for
  hand pose estimation.
\newblock In \emph{2017 IEEE Conference on Computer Vision and Pattern
  Recognition (CVPR)}. IEEE, 2017.

\bibitem[Wan et~al.(2018)Wan, Probst, Van~Gool, and Yao]{wan2018dense}
Chengde Wan, Thomas Probst, Luc Van~Gool, and Angela Yao.
\newblock Dense 3d regression for hand pose estimation.
\newblock In \emph{Proceedings of the IEEE Conference on Computer Vision and
  Pattern Recognition}, pages 5147--5156, 2018.

\bibitem[Wei et~al.(2016)Wei, Ramakrishna, Kanade, and Sheikh]{ShihEn2016}
Shih{-}En Wei, Varun Ramakrishna, Takeo Kanade, and Yaser Sheikh.
\newblock Convolutional pose machines.
\newblock In \emph{2016 IEEE Conference on Computer Vision and Pattern
  Recognition (CVPR)}, pages 4724--4732, 2016.

\bibitem[W.Yang and X.Wang(2016)]{WYang2016}
H.Li W.Yang, W.Ouyang and X.Wang.
\newblock End-to-end jearning of deformable mixture of parts and deep
  convolutional neural networks for human pose estimation.
\newblock In \emph{2016 IEEE Conference on Computer Vision and Pattern
  Recognition (CVPR)}, pages 3073--3082, 2016.

\end{thebibliography}
\end{document}